# Primary Experimental feedback on a co-manipulated Robotic system for Assisted Cervical Surgery

Seifeddine Sellemi[1], Abdelbadia Chaker[1,4], Tanguy Vendeuvre[2], Terence Essomba[3], Med Amine Laribi[1]

[1] Dept. of GMSC, Pprime Institute University of Poitiers, CNRS, ISAE-ENSMA, UPR 3346, Poitiers, France
[2] Centre Hospitalier Universitaire (CHU), Poitiers, France
[3] National Central University, Taoyuan City 32001, Taiwan
[4] National Engineering School of Sousse, University of Sousse, Sousse 4000, Tunisia.

**Abstract.** Robotic-assisted surgery has emerged as a promising approach to improve surgical ergonomics, precision, and workflow efficiency, particularly in complex procedures such as cervical spine surgery. In this study, we evaluate the performance of a collaborative robotic system designed to assist surgeons in drilling tasks by assessing its accuracy in executing predefined trajectories. A total of 14 drillings were performed by eight experienced cervical surgeons, utilizing a robotic-assisted setup aimed at ensuring stability and alignment. The primary objective of this study is to quantify the deviations in the position and orientation of the drilling tool relative to the planned trajectory, providing insights into the system's reliability and potential impact on clinical outcomes. While the primary function of robotic assistance in surgery is to enhance surgeon comfort and procedural guidance rather than solely optimizing precision, understanding the system's accuracy remains crucial for its effective integration into surgical practices part of this primary experimental feedback, the study offers an in-depth analysis of the co-manipulated robotic system's performance, focusing on the experimental setup and error evaluation methods. The findings of this study will contribute to the ongoing development of robotic-assisted cervical surgery, highlighting both its advantages and areas for improvement in achieving safer and more efficient surgical workflows.

## 1 Introduction:

The cervical spine plays a pivotal role in supporting the skull, enabling head mobility, and safeguarding the spinal cord. It comprises seven vertebrae (C1-C7), each with unique anatomical characteristics that collectively contribute to the overall mobility and stability of the neck region[1] [2].

Cervical spine surgeries are highly delicate due to the close proximity of critical neurovascular structures[3], where minor errors can lead to severe neurological deficits, vascular injuries, or impaired mobility.

The cervical spine is susceptible to various pathologies, including traumatic injuries, degenerative diseases, and tumors. In severe cases, surgical intervention is necessary to perform arthrodesis (spinal fusion), involving the fixation of vertebrae to prevent abnormal movement. This procedure often requires pedicle screw insertion[4] to stabilize the spine, a task that demands high precision due to the complex anatomy and the close proximity of neurovascular structures[5]. Given these challenges, robotic-assisted surgery has been introduced to enhance accuracy and facilitate these demanding procedures for instance the ROSA Spine system offer real-time navigation and precise guidance for the placement of pedicle screws[6]. In addition, such platform like Mazor X used to optimize the trajectory of screws during arthrodesis[7]. In the present study, , a collaborative robot is integrated into the drilling procedure of vertebrae from C2 to C7, assisting surgeons by providing guidance and improving the precision of pedicle screw placement, ultimately enhancing surgical outcomes.

This paper is structured as follows: Section 2 describes the materials and methods used in the study, including the robotic system setup and experimental protocol. Section 3 presents the results of the cervicale's drilling procedures, analyzing accuracy and deviations from the planned trajectory. Section 4 concludes the study and outlines potential future work.

## 2    Assisted robotic cervical surgery experimentation:

### 2.1    Experimental Setup

The experimental setup comprises three main components as shown in Fig.1. The first is the Franka Emika Panda collaborative robot "R", which features seven degrees of freedom. A drilling motor, equipped with a 3 mm diameter drill bit, is attached to its end effector via an intermediate support. This support secures the drilling tool to the robot, ensuring that it maintains the same orientation as the robot's end effector, as demonstrated in Figure (Figure 1). The second component is a 3D printed cervical spine model that serves as the drilling target. In our case, the model consists of six vertebrae, ranging from C2 to C7. Notably, the anatomical structure of the C2 vertebra differs from the others, as depicted in Fig.2. The model was printed using PC-ABS material, based on a CAD model generated through the segmentation of a real cadaver cervical realized by medical research partners. The third component is a navigation system, defined by a motion capture system (OptiTrack V120) and two marker clusters attached to the robot platform and to the vertebral model, respectively.

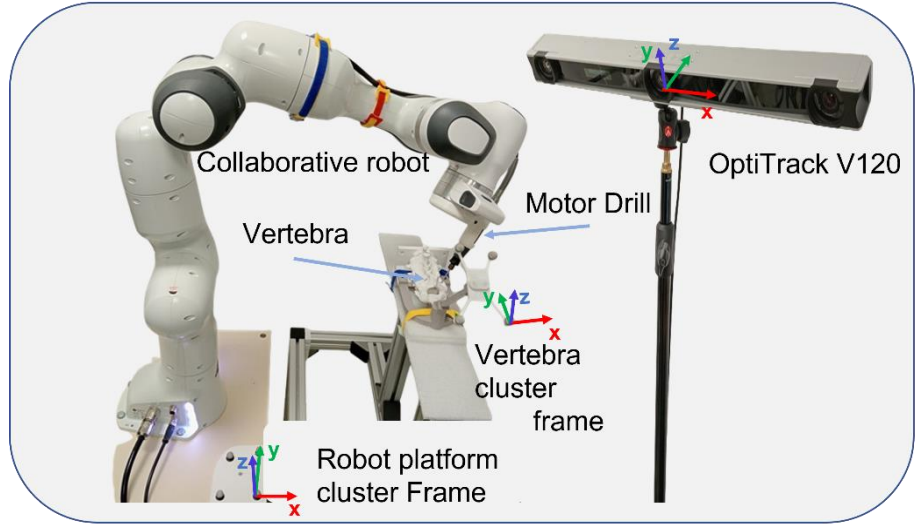

**Fig.1.** Experimental setup

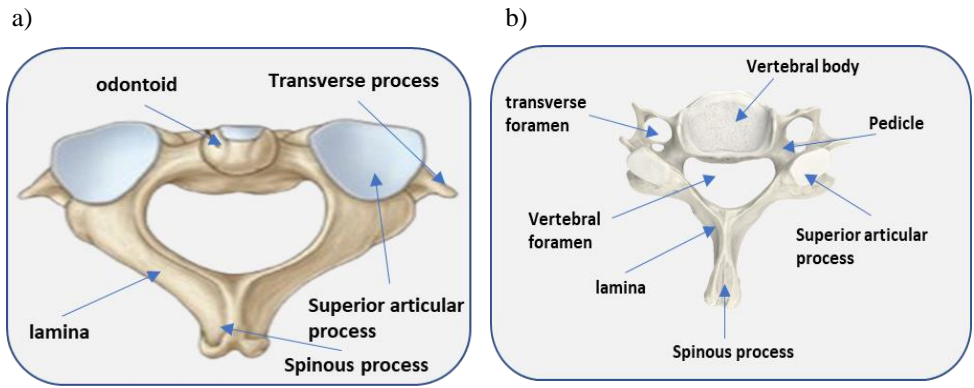

**Fig. 2.** Anatomy of the Vertebrae: a) C2 vertebra (Axis), b) C3-C7 vertebrae [2]

## 2.2 Experimental protocol

To study how well and how effectively robotic-assisted drilling works in cervical spine surgery, eight surgeons were asked to perform drilling operation on a cervical model and did sixteen drillings on the posterior side of the cervical spine. They followed a step-by-step procedure using a software interface to control the robot. the surgeon picked a vertebra (C2, C3, etc.) with the side of drilling, then the robot autonomously aligned itself with the direction desired of drilling defined by the entry point and the exit point. These points are defined by a surgeon from CHU of Poitiers in the preoperative stage. Once the robot is successfully aligned with the planned trajectory, a manual adjustment of the entry point was performed . The robot maintained its orientation, allowing the surgeon to refine the entry point and the tool's position as needed. Following confirmation of the entry point, the drilling phase commenced. All degrees of freedom of the robot were locked, except for the translational axis corresponding to the drilling direction. Prior to activation of the drilling tool, the instrument was retracted by approximately 3 cm. The drilling was then carried out to a depth of 15 mm within the vertebra. Throughout Drilling phase, the OptiTrack motion capture system and the robot's internal sensors continuously recorded positional and force data.

The recorded data include the robot's joint state, the tool tip position and the interaction forces.

## 3     Results

### 3.1     Movement performance assessment

During the experiment, the robot assisted the surgeons within the drilling operations, making sure that they were lined up with the planned directions. This assistance process is called "co-manipulation". Sixteen drilling operations were recorded and fifteen among them were considered successful with collected data. The deviation between the actual and planned directions of the drilling motor have been analyzed in order to assess how accurately surgeons followed the predefined drilling direction using the robotic assistance.

The input parameters requested to handle the drilling procedure include:

- The position and orientation of the drilling tool in the robot's frame (R).
- The vertebra coordinates in the OptiTrack camera frame (C) with frequency of 120Hz.
- The robot platform cluster coordinates in the OptiTrack camera frame (C) with frequency of 120Hz.

Because of these difference in the frequency, a temporal alignment was applied to express all data in the same reference frame, specifically the vertebra reference frame, which is defined in the OptiTrack camera frame. To achieve this, a transformation matrix is used to convert all end-effector coordinates into the vertebra reference frame (V). It is defined by the following relation:

$$\mathbf{T^{RV}} = \mathbf{T^{Rr_P}} \cdot \mathbf{T^{r_PC}} \cdot \mathbf{T^{CV}} \qquad (1)$$

With:

- $\mathbf{T^{ij}}$ : Homogeneous matrix from frame $i$ to frame $j$

- $\mathbf{T} = \begin{bmatrix} \mathbf{R} & \mathbf{P} \\ \mathbf{0} & \mathbf{1} \end{bmatrix}$, $\mathbf{R}$: Rotation matrix (3x3), $\mathbf{P}$: position vector (3x1)
- $\mathbf{R}, \mathbf{V}, \mathbf{rp}, \mathbf{C}$ are the robot, vertebra, robot platform and Camera frames, respectively

❖ **Tool position error calculation**

The position error has been computed using the Euclidean distance between the actual drilling position and its projection onto the desired drilling trajectory (Fig.4). This method ensures that the computed error accurately reflects deviations perpendicular to the direction, which are the most critical for precision in robotic-assisted drilling.

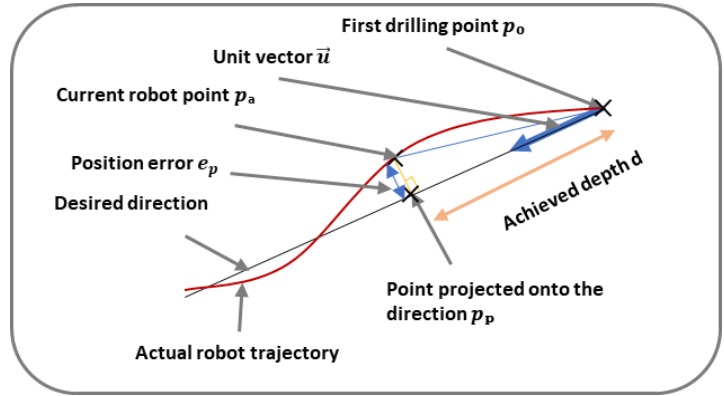

**Fig. 4.** Computation of positional error during robot-assisted drilling

The position error is calculated using the following equation:

$$\mathbf{e_p} = \|\mathbf{P_a P_0} + ((\mathbf{P_a P_0}) \cdot \mathbf{U}) \cdot \mathbf{U}\| \qquad (2)$$

Where:

- $\mathbf{P}_a$: the actual drilled point.
- $\mathbf{P}_o$: the entry point (first point in the streaming data).
- $\mathbf{U}$: the unit vector along the desired drilling trajectory.
- $(\mathbf{P}_a - \mathbf{P}_o) \cdot \mathbf{U}$: represent the projection of the positional deviation onto the drilling axis.

❖ **Tool orientation error calculation**

To evaluate the orientation deviation of the drilling tool, the error was computed with the orientation error vector based on the rotational difference between actual and target directions. The two directions were represented by the quaternion in **the vertebra reference**. The representative rotation matrixes are given as follow:

- $\mathbf{R_a}$: the matrix representing the actual tool orientation.
- $\mathbf{R_d}$: the matrix representing the desired drilling direction.

The quaternion $\mathbf{R_e}$ which represent the difference between the two orientations in the vertebrae reference and calculated in equation (Eq 3):

$$\mathbf{R}_e = \mathbf{R}_d \; \mathbf{R}_a{}^T \tag{3}$$

Where $\mathbf{R}_a{}^T$: is the transpose of $\mathbf{R}_a$

The orientation error vector $\mathbf{e_o}$ is derived from the skew-symmetric part of $\mathbf{R}_e$ and computed using the following relation (Eq 4):

$$\mathbf{e_o} = \frac{1}{2 \cdot \sin \theta} \cdot \begin{bmatrix} \mathbf{r_{32}} - \mathbf{r_{23}} \\ \mathbf{r_{13}} - \mathbf{r_{31}} \\ \mathbf{r_{21}} - \mathbf{r_{12}} \end{bmatrix} = \begin{bmatrix} \mathbf{e_x} \\ \mathbf{e_y} \\ \mathbf{e_z} \end{bmatrix} \tag{4}$$

Where:

- $\mathbf{e_o}$: Orientation error
- $\theta$: $arcos(\,1/2(\,tr(\mathbf{Re}) - 1))$ the rotation angle
- $r_{ij}$: the elements of rotation matrix $\mathbf{R}_e$

Where $\mathbf{e_x}, \mathbf{e_y}, \mathbf{e_z}$ represent the components of the orientation error vector. Each component of $\mathbf{e_o}$ indicate how much the tool deviates along the respective axes.

### 3.2 Single drilling operation analysis:

To illustrate the error evaluation process, we present the results of a representative drilling performed on the C3 in the right face and the left face (Fig. 5):

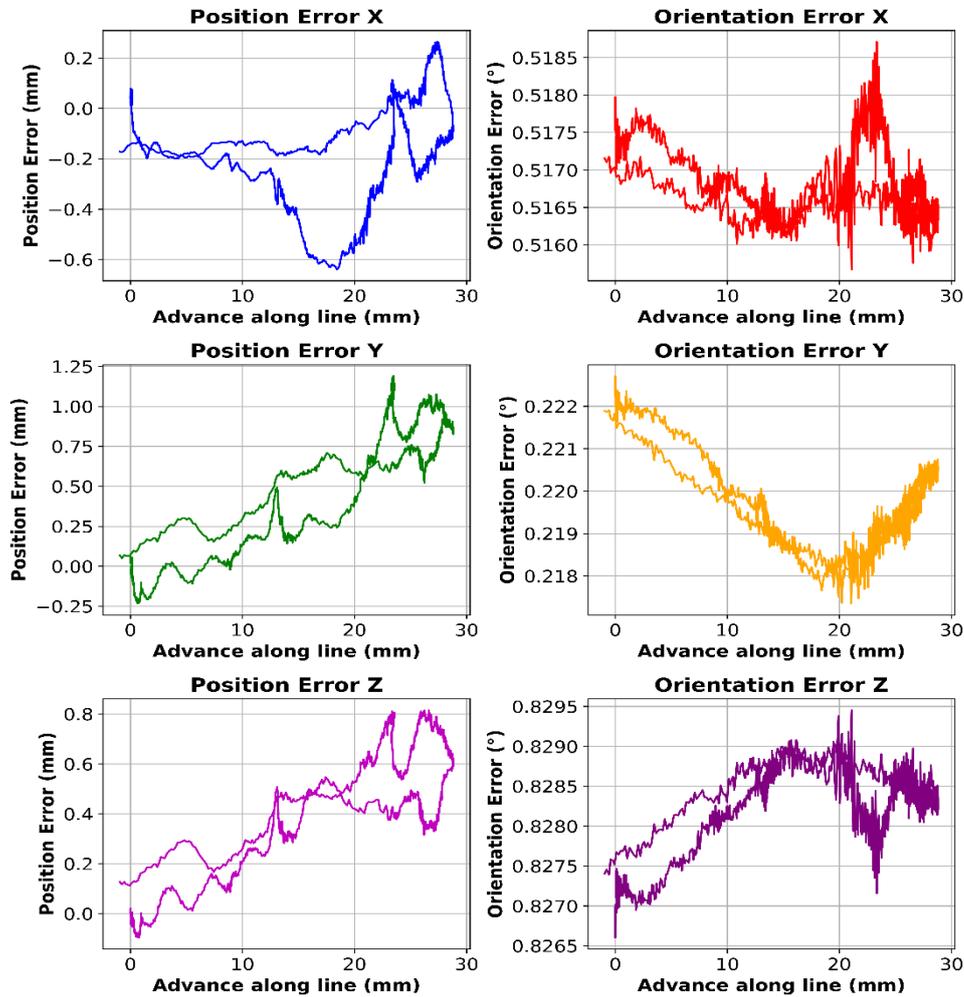

**Fig. 5:** Axis deviation of drilling (example of the C3 posterior left)

The first column of the graphs shows an example of the collected data of the position error during the drilling of the left side of the C3 vertebra. The highest error occurs along the Y-axis (1.2 mm), while the other axes remain lower. At the beginning, the error is near zero as the drill is in the air. Once the contact is established, the error gradually increases due to material resistance and forces applied by the surgeon.

The second column of the graphs illustrates orientation errors, which show instability across all axes with rapid vibrations. This is likely caused by drill vibrations and manual handling force, especially at the beginning. Despite this, the error remains below 1°, confirming that the robot helps maintain proper orientation throughout the procedure.

Based on the fifteen drilling acquisitions, a population study was conducted to evaluate the robot's suitability on the drilling process and analyze the results among different surgeons. The results obtained are depicted in Fig. 6.

a)                                                                     b)

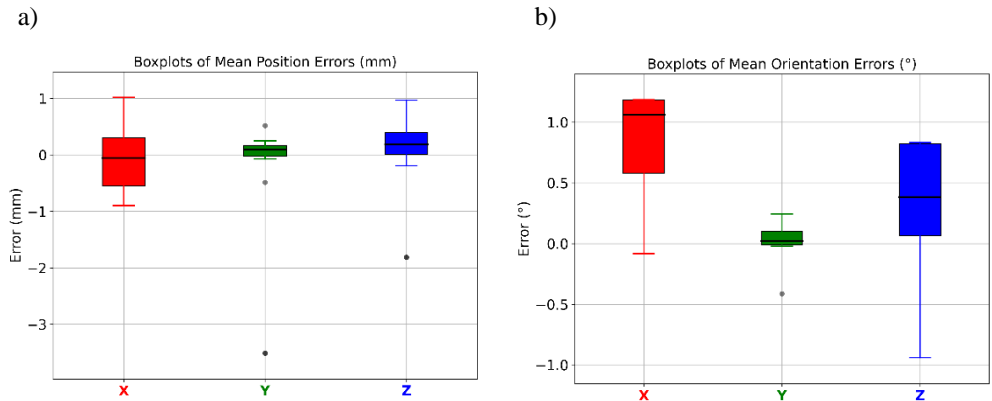

**Fig. 6:** Population Boxplot: a) error in position b) error in orientation

Fig. 6 illustrate the range of position and orientation errors for each axis. On the X-axis, the majority of the drillings (the central 50% of the data) have relatively similar errors, with a median value slightly below −0.1 mm. However, a few drillings showed much larger deviations, as indicated by the extended whiskers. This suggests a general tendency toward a small negative shift, but with some significant outliers. The Y-axis shows the smallest error range, with a median close to zero, indicating that the drilling remained well aligned. Along the Z-axis, the positional error is moderate, with a median value around 0.2 mm. The orientation error has a larger range, with a median close to 0.5 degrees, which can be explained by the presence of mechanical torques or dynamic disturbances compensated by the robot.

To visualize the results distribution over the whole population, two radar charts were generated in Fig.7. The charts display the normalized position error as well as the external forces, and the orientation error as well as the torques, respectively. The force and torque are measured by the robot at its end effector.

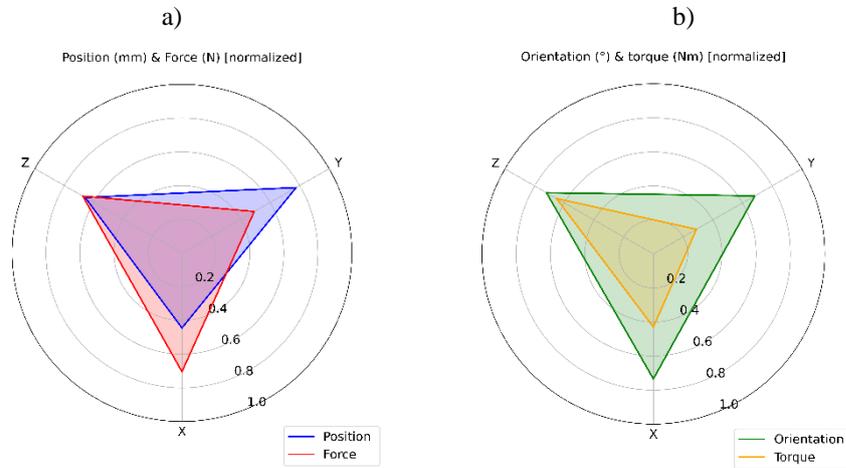

**Fig. 7**: Population radars: a) position vs. Force, b) orientation vs. Torque.

The two radar charts show that the Franka Emika Panda's behavior during drilling varies depending on the axis. Along the X-axis, the robot encounters the highest external forces, yet its positional error remains acceptable, indicating good translational stiffness. However, this same axis exhibits the largest values in orientation error and torque, pointing to low rotational stiffness. Conversely, on the Y-axis, the pattern is reversed with minimal forces related to the most significant positional error. This emphasizes high compliance (low linear stiffness) in that direction, while both angular error and torque remain acceptable. The Z-axis demonstrates intermediate values for position/force and orientation/torque, indicating balanced performance with no significant weaknesses. Overall, translation accuracy is reduced along the Y-axis, whereas rotational accuracy is most sensitive along the X-axis. This behavior is consistent with the intentionally compliant design of serial collaborative robots. It is important to note that the magnitude of these errors remains acceptable within the clinically useful range for surgical guidance. This indicates that the robotic assistance continues to provide reliable and safe assistance to the surgeon throughout the drilling procedure.

## 4 Conclusion and future work

This primary experimental study provides valuable experimental insights into the use of a collaborative robotic system for assisting in cervical surgery. Our results demonstrate that the system provides valuable help and guides the surgeons' gestures along the drilling, stabilizing the stability and the accuracy. The positional and orientation errors observed when drilling vertebrae for pedicle screw insertion are still acceptable but they need to be improved. This can be explained by some manual adjustments and material resistance contributing to minor deviations. These errors remain within acceptable clinical limits, underscoring the potential of robotic assistance in improving surgical accuracy and reducing neurovascular risks.

Future work will involve these improvements and a comparative study of manual versus robotic-assisted drilling, aiming to further quantify the benefits of robotic guidance.

## References

[1] M.-A. Rousseau, H. Pascal-Moussellard, Y. Catonné, et J.-Y. Lazennec, « Anatomie et biomécanique du rachis cervical », *Rev. Rhum.*, vol. 75, nº 8, p. 707-711, sept. 2008, doi: 10.1016/j.rhum.2008.06.001.

[2] Alizée Koszulinski, Med Amine Laribi, et Juan Sandoval, « Contribution au développement d'une plateforme d'assistance robotique pour la chirurgie cervicale », UNIVERSITE DE POITIERS, 2024.

[3] B. A. Winkelstein, R. E. McLENDON, A. Barbir, et B. S. Myers, « An anatomical investigation of the human cervical facet capsule, quantifying muscle insertion area », *J. Anat.*, vol. 198, nº 4, p. 455-461, 2001, doi: 10.1046/j.1469-7580.2001.19840455.x.

[4] A. F. Joaquim, M. L. Mudo, L. A. Tan, et K. D. Riew, « Posterior Subaxial Cervical Spine Screw Fixation: A Review of Techniques », *Glob. Spine J.*, vol. 8, nº 7, p. 751-760, oct. 2018, doi: 10.1177/2192568218759940.

[5] « (PDF) Management of Cranio-Cervical Injuries: C1-C2 posterior cervical fusion and decompression », *ResearchGate*, oct. 2024, doi: 10.1016/j.semss.2019.100782.

[6] M. Huang, T. A. Tetreault, A. Vaishnav, P. J. York, et B. N. Staub, « The current state of navigation in robotic spine surgery », *Ann. Transl. Med.*, vol. 9, nº 1, p. 86-86, janv. 2021, doi: 10.21037/atm-2020-ioi-07.

[7] M. D'Souza, J. Gendreau, A. Feng, L. H. Kim, A. L. Ho, et A. Veeravagu, « Robotic-Assisted Spine Surgery: History, Efficacy, Cost, And Future Trends », *Robot. Surg. Res. Rev.*, vol. 6, p. 9-23, nov. 2019, doi: 10.2147/RSRR.S190720.